\documentclass{article}
\usepackage{spconf,amsmath,graphicx}
\usepackage{enumitem}
\usepackage{soul}
\usepackage[dvipsnames]{xcolor}
\usepackage{tabularx}
\usepackage{multirow}
\usepackage{float}
\usepackage[T1]{fontenc}
\usepackage{hyperref}

\title{Four-in-One: A Joint Approach to Inverse Text Normalization,\\Punctuation, Capitalization, and Disfluency\\for Automatic Speech Recognition}
\name{Sharman Tan, Piyush Behre, Nick Kibre, Issac Alphonso, Shuangyu Chang}
\address{Microsoft Corporation}

\copyrightnotice{978-1-6654-7189-3/22/\$31.00~\copyright2023 IEEE}
\begin{document}
\maketitle
\begin{abstract}
Features such as punctuation, capitalization, and formatting of entities are important for readability, understanding, and natural language processing tasks.
However, Automatic Speech Recognition (ASR) systems produce spoken-form text devoid of formatting, and tagging approaches to formatting address just one or two features at a time.
In this paper, we unify spoken-to-written text conversion via a two-stage process: First, we use a single transformer tagging model to jointly produce token-level tags for inverse text normalization (ITN), punctuation, capitalization, and disfluencies. Then, we apply the tags to generate written-form text and use weighted finite state transducer (WFST) grammars to format tagged ITN entity spans.
Despite joining four models into one, our unified tagging approach matches or outperforms task-specific models across all four tasks on benchmark test sets across several domains.
\end{abstract}

\section{Introduction}
\label{sec:intro}

Automatic Speech Recognition (ASR) systems produce unstructured spoken-form text that lacks the formatting of written-form text. Converting ASR outputs into written form involves applying features such as inverse text normalization (ITN), punctuation, capitalization, and disfluency removal. ITN formats entities such as numbers, dates, times, and addresses. Disfluency removal strips the spoken-form text of interruptions such as false starts, corrections, repetitions, and filled pauses.

Spoken-to-written text conversion is critical for readability and understanding \cite{shugrina2010formatting}, as well as for accurate downstream text processing.
Prior works have emphasized the importance of well-formatted text for natural language processing (NLP) tasks including part-of-speech (POS) tagging \cite{lita2003truecasing}, named entity recognition (NER) \cite{hillard2006impact}, machine translation \cite{paulik2008sentence}, information extraction \cite{favre2008punctuating}, and summarization \cite{mrozinski2006automatic}.

The problem of spoken-to-written text conversion is complex.
Punctuation restoration requires effectively capturing long-range dependencies in text.
Techniques have evolved to do so, from $n$gram and classical machine learning approaches to recurrent neural networks and, most recently, transformers \cite{vaswani2017attention}.
Punctuation and capitalization may vary across domains, and prior works have examined legal \cite{sanchez2019sentence} and medical \cite{sunkara2020robust} texts.
ASR errors and production resource constraints pose additional challenges \cite{sunkara2020multimodal}.

ITN often involves weighted finite state transducer (WFST) grammars \cite{ebden2015kestrel} or sequence-to-sequence models \cite{mansfield2019neural}.
Punctuation, capitalization, and disfluency removal are approached as machine translation \cite{liao2020improving} or, more commonly, sequence labeling problems.
Sequence labeling tags each token in the spoken-form text to signify the desired formatting.
The translation approach is attractive as an end-to-end solution, but sequence labeling enforces structure and enables customization for domains that may only require partial formatting.
Recent works have used pre-trained transformers to jointly predict punctuation together with capitalization \cite{pappagari2021joint} and disfluency \cite{chen2020controllable}.
Prosodic features have also proven helpful for punctuation early on \cite{shriberg1997prosody} and have since been used for punctuation and disfluency detection \cite{che2016punctuation,zayats2019disfluencies}.
Despite these joint approaches, no work thus far has completely unified tagging for all four tasks.

We frame spoken-to-written text conversion as a two-stage process.
The first stage jointly tags spoken-form text for ITN, punctuation, capitalization, and disfluencies.
The second stage applies each tag sequence and outputs written-form text, employing WFST grammars for ITN and simple conversions for the remaining tasks.
To our knowledge, we are the first to jointly train a model to tag for the four tasks. We make the following key contributions:
\begin{itemize}[itemsep=-5pt]
\item We introduce a novel two-stage approach to spoken-to-written text conversion consisting of a single joint tagging model followed by a tag application stage, as described in section \ref{sec:proposed_method}
\item We define text processing pipelines for spoken- and written-form public datasets to jointly predict token-level ITN, punctuation, capitalization, and disfluency tags, as described in sections \ref{sec:text_processing_pipeline} and \ref{sec:experiments}
\item We report joint model performance on par with or exceeding task-specific models for each of the four tasks on a wide range of test sets, as described in section \ref{sec:results_and_discussion}
\end{itemize}

\section{Proposed Method}
\label{sec:proposed_method}

In this section, we describe our two-stage approach to formatting spoken-form ASR outputs. Figure \ref{fig:workflow} illustrates the end-to-end workflow of our proposed method.

\subsection{Joint labeling of ITN, punctuation, capitalization, and disfluency}

Stage 1 addresses ASR formatting as a multiple sequence labeling problem. We first tokenize the spoken-form text and then use a transformer encoder \cite{vaswani2017attention} to learn a shared representation of the input. Four task-specific classification heads -- corresponding to ITN, punctuation, capitalization, and disfluency -- predict four token-level tag sequences from the shared representation. Each classification head consists of a dropout layer followed by a fully connected layer.

We use the cross-entropy ($CE$) loss function and jointly optimize all four tasks by minimizing an evenly weighted combination of the losses as shown in

\begin{equation}
    CE_{joint} = \frac{CE_i + CE_p + CE_c + CE_d}{4}
\end{equation}

\noindent
where $CE_i$, $CE_p$, $CE_c$, and $CE_d$ are the cross-entropy loss functions for ITN, punctuation, capitalization, and disfluency, respectively.
Our task-specific experiments, described in Section \ref{sec:experiments}, optimize just the loss for the single task at hand.

\subsection{Tag application}

Stage 2 uses the four tag sequences to format the spoken-form ASR outputs as their written form.
Since the tag sequences are token-level, we convert them to word-level for tag application.

To format ITN entities, we extract each span of ITN tokens that are consecutively tagged as the same ITN entity type and span.
Then, we apply WFST grammar for that entity type to generate the written form.

ITN formatting may change the number of words in the sequence, so we preserve alignments between the original spoken-form tokens and the formatted ITN entities.
When multiple spoken-form tokens map to a single WFST output, we only apply the last punctuation tag and the first capitalization tag.
For punctuation, we append the indicated punctuation tags to the corresponding words.
For capitalization, we capitalize the first letter or entirety of words, as tagged.

To remove disfluencies, we simply remove the disfluency-tagged words from the text sequence.

Although we compare task-specific and joint models for our experiments, using four independent task-specific models in real scenarios may result in undesirable conflicts between features.
For instance, predicted punctuation may not line up with predicted beginning-of-sentence capitalization.
The joint model's shared representations encourage predictions for ITN, punctuation, capitalization, and disfluency to be consistent, avoiding such conflicts later in the tag application stage.

\begin{figure}[t]
\centering
\includegraphics[width=\linewidth]{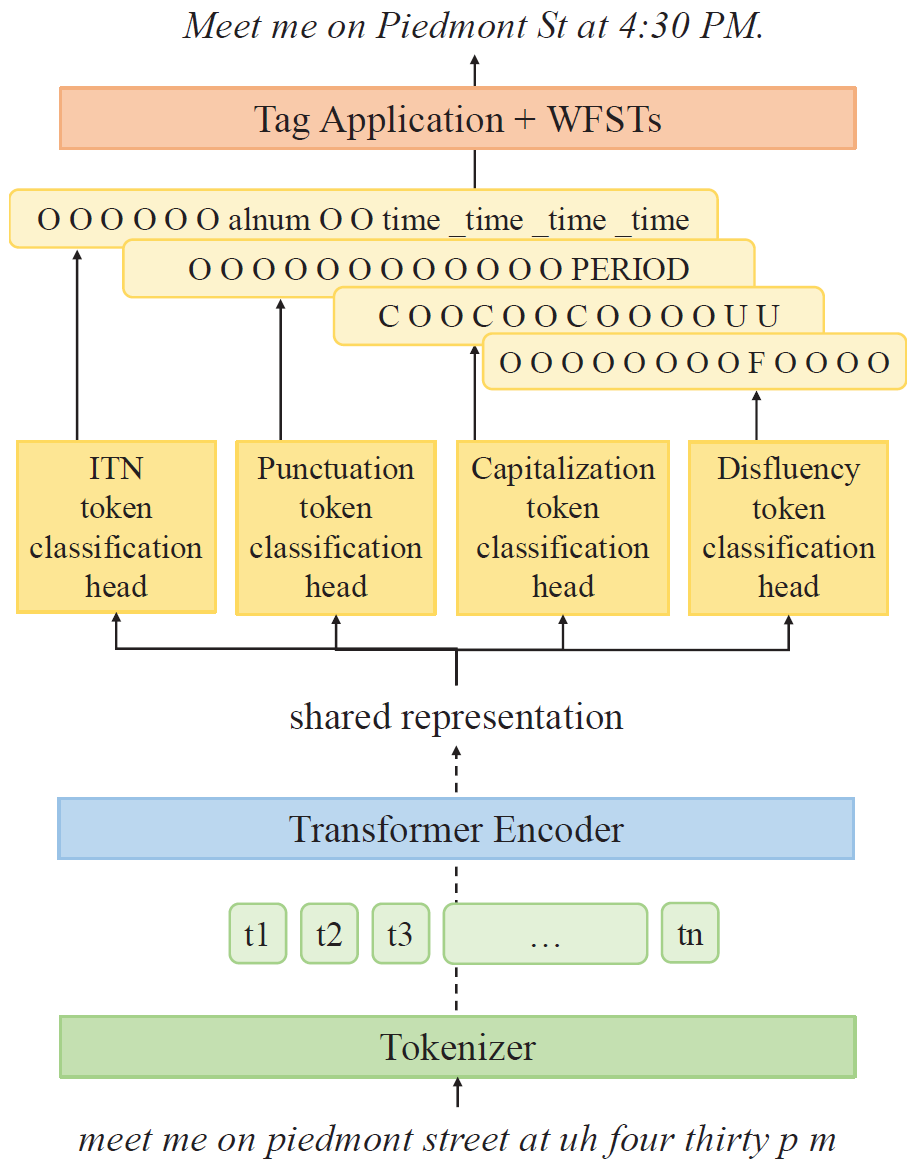}
\caption{Flow chart for joint prediction of ITN, punctuation, capitalization, and disfluency }
\label{fig:workflow}
\end{figure}

\section{Data Processing Pipeline}
\label{sec:text_processing_pipeline}

\subsection{Datasets}
We use public datasets from various domains as well as additional sets specifically targeting ITN and disfluency.
Table \ref{tab:data} shows the word count distributions by percentage among the sets.

\noindent
\textbf{OpenWebText} \cite{Gokaslan2019OpenWeb}: This dataset consists of web content extracted from URLs shared on Reddit with at least three upvotes.

\noindent
\textbf{Stack Exchange}\footnote{https://archive.org/details/stackexchange}: This dataset consists of user-contributed content on the Stack Exchange network.

\noindent
\textbf{OpenSubtitles2016} \cite{lison2016opensubtitles2016}: This dataset consists of movie and TV subtitles.
    
\noindent
\textbf{Multimodal Aligned Earnings Conference (MAEC)} \cite{li2020maec}: This dataset consists of transcribed earnings calls based on S\&P 1500 companies.

\noindent
\textbf{NPR Podcast}: This dataset consists of transcribed NPR Podcast episodes.

\noindent
\textbf{Switchboard (SWB) \& Fisher}: SWB \cite{godfrey1992switchboard} is a large multi-speaker corpus of telephone speech consisting of about 2500 conversations by 500 speakers across the United States. Fisher \cite{cieri2004fisher} is also a large conversational telephone speech corpus containing about 2000 hours of transcriptions. Neither set contains disfluency annotations.

\noindent
\textbf{Switchboard Dialog Act Corpus (SwDA)} \cite{Jurafsky-etal:1997}: This dataset consists of transcribed human-human conversational telephone speech and extends the Switchboard-1 Telephone Speech Corpus, Release 2 with turn- and utterance-level dialog-act tags.

\noindent
\textbf{Web-crawled ITN \& Conversational Disfluency}: In addition to the above sets, we also use web-crawled ITN entities (e.g., addresses, URLs, phone numbers, and numeric quantities) as well as conversational transcriptions containing disfluencies.

\begin{table}[H]
  \centering
  \begin{tabular}{|l|c|}
    \hline
    \textbf{Dataset} & \textbf{Distribution} \\
    \hline
    OpenWebText & 22.8\% \\
    Stack Exchange & 13.6\% \\
    OpenSubtitles2016 & 3.3\% \\
    MAEC & 2.9\% \\
    NPR Podcast & 0.6\% \\
    SwDA + SWB + Fisher & 0.3\% \\
    Web-crawled ITN & 56.4\% \\
    Conversational Disfluency & 0.1\% \\
    \hline
  \end{tabular}
  \caption{Data distribution by number of words per dataset}
  \label{tab:data}
\end{table}

\subsection{Data processing of written form}

Apart from SwDA, SWB, and Fisher, all of our corpora are written-form text containing ITN, punctuation, and capitalization. To jointly train a single model to predict tag sequences corresponding to ITN, punctuation, capitalization, and disfluency, we process each of our sets to contain token-level tags for each of the four tasks.

We filter and clean the datasets by preserving natural sentences or paragraphs as rows and removing characters apart from alphanumeric, punctuation, and necessary mid-word symbols such as hyphens.

To generate the spoken-form equivalents of the written-form datasets, we use WFST grammar-based text normalization.
During text normalization, we preserve alignments between written- and spoken-form ITN entities to generate ITN tags for each token.
Capitalization and punctuation tags come directly from the written form.
The written-form datasets do not contain conversational disfluencies, so we assign them all non-disfluency tags.

We reserve 10\% or at most 50 thousand rows from each set for validation and use the rest for training.

\subsection{Data processing of spoken form}

SwDA, SWB, and Fisher are spoken-form conversational transcriptions and thus do not contain ITN, capitalization, or punctuation.
Therefore, we convert the data into written form by applying a commercial formatting service.
Then, we generate ITN, capitalization, and punctuation tags using the same process as written-form datasets.

SwDA already contains dialog act annotations, so we translate these to token-level disfluency tags.
Since SWB and Fisher are not annotated for disfluencies, we follow \cite{lou2020improving} by self-training with the unannotated SWB and Fisher corpora. Specifically, we finetune an uncased BERT base model \cite{devlin-etal-2019-bert} with a top-level disfluency token classification layer on the SwDA training set and use this model to generate disfluency tags for SWB and Fisher.

We randomly split SwDA into training, validation, and test sets using a 90\%, 7\%, 3\% split, and we use all of SWB and Fisher for training. All sets are represented as byte-pair encoding (BPE) tokens for training and evaluation.

\subsection{Tag classes}
\noindent
\textbf{ITN}: We tag each token as one of 5 entity types (alphanumeric, numeric, ordinal, money, time) or `O' representing non-ITN.
Since we apply WFST grammars on each spoken-form ITN entity span, we signify each ITN entity span by tagging the first token as the entity tag and prepending an underscore (`\_') character to the remaining tags in the span. For example, ``four thirty p m" is tagged as ``time \_time \_time \_time" as illustrated in Figure \ref{fig:workflow}.

\noindent
\textbf{Punctuation}: We define 4 tag categories: comma, period, question mark, and `O' for no punctuation.
Each punctuation tag represents the punctuation that appears appended to the corresponding token of text.

\noindent
\textbf{Capitalization}: We define 3 tag categories: all uppercase (`U'), capitalize only the first letter (`C'), and all lowercase (`O').

\noindent
\textbf{Disfluency}: Following SwDA annotations \cite{Jurafsky-etal:1997, Shriberg-etal:1998, Stolcke-etal:2000}, we define 7 tag categories: correction of repetition (`C\_RT'), reparandum repetition (`R\_RT'), correction (`C'), reparandum (`R'), filler word (`F'), all other disfluency (`D'), and non-disfluency (`O').

\section{Experiments}
\label{sec:experiments}

\subsection{Test sets}

We evaluate our task-specific and joint models on public and private test sets.
\newline
\newline
\noindent
\textbf{IWSLT 2011 TED Talks}\footnote{http://iwslt2011.org/doku.php?id=06\_evaluation}: The IWSLT 2011 ASR and reference test sets each consists of one continuous stream of text with ground truth punctuation. Because our models expect sentence- or paragraph-level input, we convert the ASR and reference sets into sentences and then form paragraphs of at most 200 words, keeping sentences intact. This results in 322 paragraphs. We generate spoken-form inputs for our model evaluation using text normalization.

\noindent
\textbf{DeepMind Q\&A CNN and DailyMail stories} \cite{hermann2015teaching}: Each of these sets consists of 10,000 written-form paragraphs extracted from randomly selected stories. Each paragraph is well-formed, at least 3 sentences long, and does not contain extraneous symbols such as quotation marks or parentheses. We generate spoken-form inputs for our model evaluation using text normalization.

\noindent
\textbf{NPR Podcasts}\footnote{https://www.npr.org}: This set consists of 76 transcribed NPR podcast episodes. For ASR evaluation, we run a commercial ASR service\footnote{Azure Speech Recognition} on the original audio files to generate corresponding spoken-form ASR outputs.

\noindent
\textbf{Google TN} \cite{sproat2016rnn}: This set consists of written texts aligned to their normalized spoken forms, where the normalizations were generated using an existing text normalization component of a text-to-speech system. We keep only well-formatted sentences over 3 words long, contain ITN entities, and do not contain extraneous symbols such as quotation marks or parentheses. We use the top 10,000 remaining sentences for testing.

\noindent
\textbf{SwDA} \cite{Jurafsky-etal:1997}: We randomly selected 6,445 examples from SwDA as our disfluency test set.

\noindent
\textbf{Conversational Disfluency}: This set consists of 1,646 lines of conversational spoken-form text in which 23.6\% of tokens are disfluencies.

\noindent
\textbf{Dictation}: This internal set consists of 100 utterances of long-form dictation ASR outputs and human labeled transcriptions.

\noindent
\textbf{Voicemails}: This internal set consists of 400 voicemail ASR outputs and human labeled transcriptions.

\noindent
\textbf{Web-crawled ITN test set}: This manually curated set of ITN entities is designed to comprehensively evaluate model performance on each ITN entity category.
\subsection{Experimental Setup}
We hypothesize that the joint model will perform at least on par with task-specific models, as the joint model may better leverage synergies in language between ITN, punctuation, capitalization, and disfluencies.

To test our hypothesis, we train one task-specific model for each of the four tasks and one joint model.
To directly compare task-specific and joint models, we train the models with the same training and validation sets, using the same model architecture and hyperparameters. Each model is trained to convergence.

Both task-specific and joint models are 12-layer transformers with 16 attention heads, 1024-dimension word embeddings, 4096-dimension fully connected layers, and 8-dimension projection layers between the transformer encoder and the decoder that maps to the tag classes.
The joint model contains four parallel projection layers, one for each of the tasks.
From here, we refer to the task-specific models collectively as TASK and the joint model as JOINT.

Because the joint model only has a quarter of the parameters of the four task-specific models combined, we also compare JOINT to smaller task-specific models TASK-SMALL with 512-dimension word embeddings and 2048-dimension fully connected layers.
TASK-SMALL in total trains about 204 million parameters, compared to JOINT which trains about 171 million parameters.

Disfluency-specific training data makes up just 0.4\% of the total training data, so we also train a task-specific model on just the conversational datasets (SwDA, SWB, Fisher). We refer to this disfluency detection model as TASK-DISF.

\begin{table*}[t]
  \centering
  \begin{tabular}{|c|c|c|ccc|ccc|ccc|ccc|}
    \hline
    & \multirow{2}{*}{\textbf{Test Set}} & \multirow{2}{*}{\textbf{Model}} &
    \multicolumn{3}{c|}{\textbf{COMMA}} &
    \multicolumn{3}{c|}{\textbf{PERIOD}} &
    \multicolumn{3}{c|}{\textbf{Q-MARK}} &
    \multicolumn{3}{c|}{\textbf{OVERALL}} \\
    \cline{4-15}
    & & & P & R & F$_1$ &
    P & R & F$_1$ &
    P & R & F$_1$ &
    P & R & F$_1$ \\
    \hline
    \multirow{3}{*}{Ref.} &
    \multirow{3}{*}{\textit{CNN Stories}} &
    TASK-SMALL &
    84 & 80 & 82 &
    90 & 83 & 86 &
    85 & 83 & 84 &
    86 & 81 & 84 \\
    & & TASK &
    84 & \textbf{82} & \textbf{83} &
    \textbf{91} & 84 & 87 &
    86 & \textbf{85} & 85 &
    \textbf{87} & \textbf{83} & \textbf{85} \\
    & & \textbf{JOINT} &
    \textbf{84} & 81 & 82 &
    90 & \textbf{84} & \textbf{87} &
    \textbf{86} & 83 & \textbf{85} &
    86 & 82 & 84 \\
    \hline
    \multirow{3}{*}{Ref.} &
    \multirow{3}{*}{\textit{DailyMail Stories}} &
    TASK-SMALL &
    76 & 79 & 77 &
    90 & 88 & 89 &
    82 & 71 & 76 &
    82 & 83 & 82 \\
    & & TASK &
    77 & \textbf{80} & \textbf{79} &
    \textbf{92} & \textbf{90} & \textbf{91} &
    \textbf{88} & \textbf{78} & \textbf{83} &
    84 & 85 & 84 \\
    & & \textbf{JOINT} &
    \textbf{77} & 79 & 78 &
    91 & 89 & 90 &
    84 & 74 & 78 &
    \textbf{83} & \textbf{84} & \textbf{83} \\
    \hline
    \multirow{3}{*}{ASR} &
    \multirow{3}{*}{\textit{IWSLT 2011 TED}} &
    TASK-SMALL &
    66 & 31 & 43 &
    73 & 68 & 71 &
    59 & 42 & 49 &
    69 & 49 & 56 \\
    & & TASK &
    68 & \textbf{33} & \textbf{44} &
    75 & \textbf{68} & \textbf{72} &
    56 & \textbf{45} & \textbf{50} &
    71 & \textbf{50} & \textbf{57} \\
    & & \textbf{JOINT} &
    \textbf{70} & 20 & 31 &
    \textbf{75} & 67 & 71 &
    \textbf{80} & 26 & 39 &
    \textbf{73} & 42 & 50 \\
    \hline
    \multirow{3}{*}{Ref.} &
    \multirow{3}{*}{\textit{IWSLT 2011 TED}} &
    TASK-SMALL &
    78 & 61 & 69 &
    81 & 88 & 85 &
    80 & 85 & 82 &
    79 & 74 & 77 \\
    & & TASK &
    79 & \textbf{67} & \textbf{72} &
    \textbf{84} & \textbf{88} & \textbf{86} &
    71 & \textbf{90} & 80 &
    \textbf{81} & \textbf{77} & \textbf{79} \\
    & & \textbf{JOINT} &
    \textbf{79} & 63 & 70 &
    82 & 87 & 85 &
    \textbf{80} & 85 & \textbf{82} &
    80 & 75 & 77 \\
    \hline
    \multirow{3}{*}{ASR} &
    \multirow{3}{*}{\textit{NPR Podcasts}} &
    TASK-SMALL &
    71 & 60 & 65 &
    83 & 77 & 80 &
    80 & 68 & 74 &
    77 & 69 & 73 \\
    & & TASK &
    71 & \textbf{62} & \textbf{67} &
    \textbf{84} & \textbf{78} & \textbf{81} &
    80 & 68 & 74 &
    78 & \textbf{70} & \textbf{74} \\
    & & \textbf{JOINT} &
    \textbf{71} & 60 & 65 &
    83 & 77 & 80 &
    \textbf{82} & \textbf{69} & \textbf{75} &
    \textbf{78} & 69 & 73 \\
    \hline
    \multirow{3}{*}{ASR} &
    \multirow{3}{*}{\textit{Dictation}} &
    TASK-SMALL &
    69 & 54 & 61 &
    72 & 78 & 75 &
    48 & 94 & \textbf{64} &
    70 & 65 & 67 \\
    & & TASK &
    70 & \textbf{57} & \textbf{63} &
    73 & 79 & 76 &
    44 & \textbf{94} & 60 &
    71 & 67 & \textbf{69} \\
    & & \textbf{JOINT} &
    \textbf{70} & 56 & 62 &
    \textbf{73} & \textbf{80} & \textbf{76} &
   \textbf{ 50} & 81 & 62 &
    \textbf{71} & \textbf{67} & 68 \\
    \hline
    \multirow{3}{*}{Ref.} &
    \multirow{3}{*}{\textit{Dictation}} &
    TASK-SMALL &
    73 & 59 & 65 &
    82 & 76 & 79 &
    71 & 92 & 80 &
    77 & 66 & 71 \\
    & & TASK &
    73 & 61 & 66 &
    83 & \textbf{78} & 80 &
    65 & 100 & 79 &
    77 & 68 & 72 \\
    & & \textbf{JOINT} &
    \textbf{73} & \textbf{61} & \textbf{66} &
    \textbf{85} & 77 & \textbf{81} &
    \textbf{72} & \textbf{100} & \textbf{84} &
    \textbf{78} & \textbf{68} & \textbf{72} \\
    \hline
  \end{tabular}
  \caption{Punctuation results}
  \label{tab:punc}
\end{table*}

\section{Results}
\label{sec:results_and_discussion}
We measure performance of each task on test sets using word-level precision (P), recall (R), and F$_1$ scores.

Table \ref{tab:punc} presents punctuation results on all relevant test sets.
JOINT performance is consistently on par with TASK, achieving similar F$_1$ scores across sets and tag classes.
In the period and question mark categories, JOINT outperforms both task-specific models on several sets.
Across the test sets, JOINT mostly maintains or improves upon TASK precision, sometimes sacrificing some points in recall.
In customer scenarios such as long-form dictation, a tilt towards precision is often preferable to avoid over-punctuation.

From Tables \ref{tab:itn} and \ref{tab:cap}, we see that JOINT almost always achieves the best capitalization and ITN F$_1$ across the board, matching or outperforming both task-specific models.
Punctuation improvements in detecting sentence boundaries contributed to some gains in beginning-of-sentence capitalization.
Jointly training to predict ITN entities helped JOINT to correctly capitalize difficult entities such as addresses and alphanumeric codes.

Table \ref{tab:disf} shows results from evaluating the models on the SwDA test set and an internal conversational test set containing disfluencies.
On both sets, we see that JOINT significantly outperforms TASK, achieving 44\% higher F$_1$ on SwDA and 69\% higher F$_1$ on the Disfluency test. JOINT matches the performance of TASK-DISF, which is trained only on SwD, SWB, and Fisher corpora.

This disparity between TASK and TASK-DISF performance reflects TASK's training skew towards fluent rather than disfluent data; understandably, TASK achieves low recall.
Even though we train JOINT on the same imbalanced data as TASK, we see that JOINT achieves performance on par with TASK-DISF.
While the F$_1$ scores of TASK-DISF and JOINT suggest the two perform similarly, qualitative evaluation on out-of-domain test cases reveals important differences.
\begin{table}[H]
  \centering
  \begin{tabular}{|c|c|c|ccc|ccc|ccc|ccc|}
    \hline
    & \multirow{2}{*}{\textbf{Test Set}} & \multirow{2}{*}{\textbf{Model}} & \multicolumn{3}{c|}{\textbf{ITN}} \\
    \cline{4-6}
    & & & P & R & F$_1$ \\
    \hline
    \multirow{3}{*}{Ref.} &
    \multirow{3}{1.7cm}{\centering \textit{CNN Stories}} &
    TASK-SMALL &
    88 & 87 & 88 \\
    & & TASK &
    88 & 87 & 88 \\
    & & \textbf{JOINT} &
    \textbf{89} & \textbf{87} & \textbf{88} \\
    \hline
    \multirow{3}{*}{Ref.} &
    \multirow{3}{1.7cm}{\centering \textit{DailyMail Stories}} &
    TASK-SMALL &
    84 & 84 & 84 \\
    & & TASK &
    84 & 84 & 84 \\
    & & \textbf{JOINT} &
    \textbf{85} & \textbf{84} & \textbf{85} \\
    \hline
    \multirow{3}{*}{ASR} &
    \multirow{3}{1.7cm}{\centering \textit{NPR Podcasts}} &
    TASK-SMALL &
    76 & 58 & 66 \\
    & & TASK &
    77 & 59 & 66 \\
    & & \textbf{JOINT} &
    \textbf{77} & \textbf{59} & \textbf{67} \\
    \hline
    \multirow{3}{*}{Ref.} &
    \multirow{3}{1.7cm}{\centering \textit{Wikipedia}} &
    TASK-SMALL &
    \textbf{65} & 69 & \textbf{67} \\
    & & TASK &
    63 & 68 & 66 \\
    & & \textbf{JOINT} &
    64 & \textbf{69} & 66 \\
    \hline
    \multirow{3}{*}{ASR} &
    \multirow{3}{1.7cm}{\centering \textit{Dictation}} &
    TASK-SMALL &
    75 & 59 & 66 \\
    & & TASK &
    74 & 58 & 65 \\
    & & \textbf{JOINT} &
    \textbf{76} & \textbf{60} & \textbf{67} \\
    \hline
    \multirow{3}{*}{Ref.} &
    \multirow{3}{1.7cm}{\centering \textit{Dictation}} &
    TASK-SMALL &
    84 & 62 & 72 \\
    & & TASK &
    83 & 62 & 71 \\
    & & \textbf{JOINT} &
    \textbf{84} & \textbf{63} & \textbf{72} \\
    \hline
    \multirow{3}{*}{Ref.} &
    \multirow{3}{1.9cm}{\centering \textit{Web-crawled ITN}} &
    TASK-SMALL &
    82 & 76 & 79 \\
    & & TASK &
    85 & 75 & 78 \\
    & & \textbf{JOINT} &
    \textbf{82} & \textbf{76} & \textbf{79} \\
    \hline
  \end{tabular}
  \caption{ITN results}
  \label{tab:itn}
\end{table}
\begin{table}[H]
  \centering
  \begin{tabular}{|c|c|c|ccc|}
    \hline
    & \multirow{2}{*}{\textbf{Test Set}} & \multirow{2}{*}{\textbf{Model}} & \multicolumn{3}{c|}{\textbf{DISFLUENCY}} \\
    \cline{4-6}
    & & & P & R & F$_1$ \\
    \hline
    \multirow{3}{*}{Ref.} &
    \multirow{3}{1.5cm}{\centering \textit{SwDA}} &
    TASK-DISF &
    \textbf{95} & 84 & 89\\
    & & TASK &
    89 & 47 & 62\\
    & & \textbf{JOINT} &
    94 & \textbf{85} & \textbf{89}\\
    \hline
    \multirow{3}{*}{Ref.} &
    \multirow{3}{1.5cm}{\centering \textit{Conv. Disfluency}} &
    TASK-DISF &
    \textbf{78} & \textbf{44} & \textbf{56}\\
    & & TASK &
    72 & 20 & 32\\
    & & \textbf{JOINT} &
    76 & 42 & 54\\
    \hline
  \end{tabular}
  \caption{Disfluency results}
  \label{tab:disf}
\end{table}

In the example voicemail in Table \ref{tab:qual}, TASK-DISF incorrectly tags ``oh'' as disfluency, while JOINT correctly detects that it is part of a phone number.
In real customer scenarios, tagging and removing ``oh'' as disfluency would completely alter the phone number and render it unusable.
By jointly training on a wide range of non-disfluency data, we generalize well to other domains and avoid critical false positives in production environments.

\begin{table*}[ht!]
  \centering
  \begin{tabular}{|c|c|c|ccc|ccc|ccc|ccc|}
    \hline
    & \multirow{2}{*}{\textbf{Test Set}} & \multirow{2}{*}{\textbf{Model}} &
    \multicolumn{3}{c|}{\textbf{UPPERCASE}} &
    \multicolumn{3}{c|}{\textbf{CAPITAL}} &
    \multicolumn{3}{c|}{\textbf{SINGLE-CASE}} &
    \multicolumn{3}{c|}{\textbf{OVERALL}} \\
    \cline{4-15}
    & & & P & R & F$_1$ &
    P & R & F$_1$ &
    P & R & F$_1$ &
    P & R & F$_1$ \\
    \hline
    \multirow{3}{*}{Ref.} &
    \multirow{3}{*}{\textit{CNN Stories}} &
    TASK-SMALL &
    38 & 83 & 52 &
    94 & 92 & 93 &
    97 & 49 & 66 &
    93 & 84 & 87 \\
    & & TASK &
    38 & 83 & 52 &
    94 & 93 & 94 &
    97 & 49 & 66 &
    93 & 85 & 88 \\
    & & \textbf{JOINT} &
    \textbf{38} & \textbf{83} & \textbf{53} &
    \textbf{95} & \textbf{93} & \textbf{94} &
    \textbf{97} & \textbf{49} & \textbf{66} &
    \textbf{93} & \textbf{85} & \textbf{88} \\
    \hline
    \multirow{3}{*}{Ref.} &
    \multirow{3}{*}{\textit{DailyMail Stories}} &
    TASK-SMALL &
    82 & 92 & 87 &
    94 & 93 & 93 &
    91 & 77 & 84 &
    93 & 92 & 92 \\
    & & TASK &
    81 & \textbf{93} & 87 &
    94 & 94 & 94 &
    92 & 77 & 84 &
    93 & 93 & 93 \\
    & & \textbf{JOINT} &
    \textbf{82} & 92 & \textbf{87} &
    \textbf{95} & \textbf{94} & \textbf{94} &
    \textbf{92} & \textbf{77} & \textbf{84} &
    \textbf{94} & \textbf{93} & \textbf{93} \\
    \hline
    \multirow{3}{*}{ASR} &
    \multirow{3}{*}{\textit{NPR Podcasts}} &
    TASK-SMALL &
    80 & 68 & 74 &
    88 & 83 & 86 &
    93 & \textbf{84} & \textbf{88} &
    88 & 82 & 86 \\
    & & TASK &
    83 & 69 & 75 &
    89 & 84 & 86 &
    90 & 81 & 85 &
    89 & 83 & 85 \\
    & & \textbf{JOINT} &
    \textbf{83} & \textbf{69} & \textbf{75} &
    \textbf{89} & \textbf{84} & \textbf{86} &
    \textbf{93} & 82 & 87 &
    \textbf{89} & \textbf{83} & \textbf{86} \\
    \hline
    \multirow{3}{*}{ASR} &
    \multirow{3}{*}{\textit{Dictation}} &
    TASK-SMALL &
    75 & 74 & 74 &
    79 & 82 & 81 &
    60 & 72 & 65 &
    78 & 81 & 80 \\
    & & TASK &
    73 & 77 & 75 &
    80 & 83 & 81 &
    54 & 64 & 59 &
    78 & 82 & 79 \\
    & & \textbf{JOINT} &
    \textbf{75} & \textbf{79} & \textbf{77} &
    \textbf{80} & \textbf{83} & \textbf{82} &
    \textbf{63} & \textbf{77} & \textbf{69} &
    \textbf{79} & \textbf{82} & \textbf{81} \\
    \hline
    \multirow{3}{*}{Ref.} &
    \multirow{3}{*}{\textit{Dictation}} &
    TASK-SMALL &
    77 & 88 & 82 &
    88 & 83 & 85 &
    72 & 53 & 61 &
    86 & 82 & 84 \\
    & & TASK &
    75 & 88 & 81 &
    87 & \textbf{85} & 86 &
    72 & 53 & 61 &
    85 & \textbf{84} & 84 \\
    & & \textbf{JOINT} &
    \textbf{78} & \textbf{88} & \textbf{83} &
    \textbf{88} & 84 & \textbf{86} &
    \textbf{73} & \textbf{56} & \textbf{63} &
    \textbf{86} & 83 & \textbf{85} \\
    \hline
  \end{tabular}
  \caption{Capitalization results. Uppercase refers to words longer than 1 letter that are uppercase, Capital refers to words with only first letter capitalized, and Single-case refers to 1-letter words that are uppercase.}
  \label{tab:cap}
\end{table*}
\begin{table*}[ht!]
  \centering
  \begin{tabular}{ll}
    \textbf{Spoken-form Input} & please call me back at eight \textbf{oh} five six seven zero zero four two three \\
    \textbf{TASK(-DISF) Output} & Please call me back at 856700423. \\
    \textbf{JOINT Output} & Please call me back at 8\textbf{0}5-670-0423.\\
  \end{tabular}
  \caption{Example in which TASK-DISF mistakes `oh' as disfluency, while JOINT correctly formats the phone number}
  \label{tab:qual}
\end{table*}

Our results indicate that jointly labeling ITN, punctuation, capitalization, and disfluency achieves performance on par with four equivalent task-specific models, despite a 75\% reduction in parameters.
Even when TASK outperforms JOINT, JOINT still achieves equal or better F$_1$ compared to TASK-SMALL which trains 33 million more total parameters.
Therefore, our approach not only maintains or improves task-specific performance but also significantly cuts training and runtime costs for formatting in real scenarios.

\section{Related Work}
\label{sec:related_work}

ITN formatting approaches typically involve rule-based systems such as WFST grammars \cite{NEUBIG2012349} and sequence-to-sequence models \cite{mansfield2019neural}.
WFST grammars promise accurate results \cite{ebden2015kestrel}, and recent work has explored combining sequence-to-sequence models and WFST grammars for production \cite{sunkara2021neural}.

While punctuation, capitalization, and disfluency detection tasks have long been addressed as sequence labeling problems, the techniques to solve them have evolved. Unigram and $n$gram language models are straightforward but suffer from limited knowledge of surrounding context and lack of scalability as $n$ grows large \cite{lita2003truecasing,gravano2009restoring}.
Classical machine learning techniques for punctuation and disfluency detection use hidden markov models (HMMs) \cite{liu2006enriching}, maximum entropy models \cite{huang2002maximum}, and conditional random fields (CRFs) \cite{lu2010better}.
However, these approaches require manual feature engineering and are cumbersome to train.

These classical techniques gave way to deep neural network approaches such as recurrent neural networks (RNNs), which are easier to train and able to learn more complex features.
RNNs and especially long-short term memory (LSTM) models, sometimes combined with CRF layers, have effectively leveraged surrounding context to predict punctuation \cite{che2016punctuation,yi2017distilling}, capitalization \cite{zhang2022capitalization}, and disfluency detection \cite{zayats2016disfluency,wang2016neural}.
Most recently, pre-trained transformers have dominated the state-of-the-art in spoken-to-written text conversion.

Multiple sequence labeling approaches have addressed jointly training punctuation with capitalization \cite{pappagari2021joint} and disfluency detection \cite{chen2020controllable,liu2006enriching}.
However, we are the first to jointly label all four key components of ASR formatting.
Joint learning not only takes advantage of natural correlations in language, but also drastically reduces latency and memory costs in production.
Furthermore, our hybrid ITN approach has the best of both worlds -- better context-based detection of ITN entities and highly accurate WFST grammars.

\section{Conclusion}
\label{sec:conclusion}
In this paper, we introduced a four-in-one approach to ITN, punctuation, capitalization, and disfluency removal for spoken-to-written text conversion. Our joint transformer tagging model, trained from scratch on spoken- and written-form data, matches or exceeds the performance of task-specific models on all four tasks across test domains -- despite training a fraction of the parameters. Our approach can be extended to jointly tag for POS, NER, and other NLP tasks. Future work will explore prosodic cues and multilingual modeling.

\section{Acknowledgements}
\label{sec:acknowledgements}
We thank Yashesh Gaur and our colleagues at Microsoft for their work on streaming ITN using sequence labeling and WFST grammars, which we built upon in our approach.

\vfill\pagebreak
\clearpage

\bibliographystyle{IEEEbib}
\bibliography{refs}

\end{document}